\documentclass[letterpaper, 10 pt, journal, twoside]{IEEEtran}
\usepackage{bm} 

\usepackage{graphics} 
\graphicspath{{figures/}}
\usepackage{epsfig} 
\usepackage{mathptmx} 
\DeclareMathAlphabet{\mathcal}{OMS}{cmsy}{m}{n}
\DeclareSymbolFont{largesymbols}{OMX}{cmex}{m}{n}
\usepackage{amsmath} 
\usepackage{amssymb}  

\usepackage{cite} 

\usepackage[colorlinks,linkcolor=red]{hyperref} %
\usepackage{color}
\usepackage{multirow}
\usepackage{soul}

\usepackage{balance} %
\ifCLASSOPTIONcompsoc
 \usepackage[caption=false,font=normalsize,labelfont=sf,textfont=sf]{subfig}
\else
 \usepackage[caption=false,font=footnotesize]{subfig}
\fi
\begin{document}

\title{Edge-based Monocular Thermal-Inertial Odometry in Visually Degraded Environments}

\author{Yu~Wang, Haoyao~Chen*,~\IEEEmembership{Member,~IEEE}, Yufeng~Liu, Shiwu~Zhang,~\IEEEmembership{Member,~IEEE}
  \thanks{This work was supported in part by the National Natural Science Foundation of China under Grant U1713206 and U21A20119. (Corresponding author: Haoyao Chen.)}
  \thanks{Y. Wang, H. Chen* ,and Y. Liu are with the School of Mechanical Engineering and Automation, Harbin Institute of Technology Shenzhen, P.R. China, e-mail: hychen5@hit.edu.cn.}
  \thanks{S. Zhang is with Department of Precision Mechinery and Precision Instrumentation, University of Science and Technology of China.}
}
\maketitle

\begin{abstract}
	State estimation in complex illumination environments based on 
	conventional visual-inertial odometry is a challenging task due to 
	the severe visual degradation of the visual camera. The thermal infrared 
	camera is capable of all-day time and is less affected 
	by illumination variation. However, most existing visual data 
	association algorithms are incompatible because the 
	thermal infrared data contains large noise and low contrast. Motivated 
	by the phenomenon that thermal radiation varies most significantly 
	at the edges of objects, the study proposes an ETIO, which is the first 
	edge-based monocular thermal-inertial odometry for robust localization in visually 
	degraded environments. Instead of the raw image, 
	we utilize the binarized image from edge extraction for pose estimation 
	to overcome the poor thermal infrared image quality. Then, an adaptive 
	feature tracking strategy ADT-KLT is developed for robust data 
	association based on limited edge information and its distance distribution. 
	Finally, a pose graph optimization performs real-time estimation over a sliding window 
	of recent states by combining IMU pre-integration with reprojection 
	error of all edge feature observations. We evaluated the performance 
	of the proposed system on public datasets and real-world experiments 
	and compared it against state-of-the-art methods. The proposed 
	ETIO was verified with the ability to enable accurate and robust 
	localization all-day time.
\end{abstract}

\begin{IEEEkeywords}
Thermal-Inertial Odometry, Edge information, Visual Degradation.
\end{IEEEkeywords}

%
\IEEEpeerreviewmaketitle

\section{INTRODUCTION}
%
%
%
%
\IEEEPARstart{A}{ccurate} and robust state estimation in GNSS-denied environments is 
an active research field due to its wide applications in simultaneous localization and 
mapping (SLAM), 3D reconstruction, and active exploration. The sensor suit consisting of 
a monocular camera and IMU, which provides complementary information, 
is the minimum solution for recovering the metric six degrees-of-freedom (DOF)\cite{qin2018vins}. 
Considering that both camera and IMU are light-weight and low-cost, monocular 
visual-inertial odometry (VIO) is a common solution for localization and 
navigation\cite{von2022dm}. Existing VIO frameworks have been mature in stable 
environments. However, the environments in disaster areas are uncertain 
and prone to extreme light distribution, dynamic illumination variation, 
or visual obscurants such as dust, fog, and smoke\cite{khattak2020keyframe}. 
Such visual degradation always reduces the reliability of VIO estimation 
solutions. 

Thermal infrared camera, operating in the longwave infrared spectrum and 
capturing thermal-radiometric information, has attracted more 
attention in recent years. Compared with the visual camera, thermal infrared 
cameras exhibit evident advantages when applied to disaster areas for their 
all-day perceptual capability\cite{banuls2020object}. However, using thermal 
infrared cameras directly to existing VIO frameworks is challenging for the 
following reasons. First, the captured image data are typically low contrast\cite{shin2020dvl}.
Second, many vision-observable information-rich textures, such as colors and streaks, 
are lost in thermal images due to the indistinguishability of thermal radiation from 
surrounding regions. Lastly, nonuniformity correction (NUC) or flat‐field 
correction (FFC) is performed during thermal infrared camera operation 
to eliminate accumulated nonzero‐mean noise\cite{2004Nonuniformity}. 
Such blackout not only introduces periods of data interruption 
but may also significantly change image appearance between consecutive 
frames.

The current thermal-inertial odometry (TIO) solutions are mainly improved from normal VIO. 
Feature-based thermal odometry that requires special contrast 
enhancement on infrared images for feature extraction was 
developed \cite{Beauvisage2016Multi,Poujol2016A,article}. 
However, preprocessing will induce additional noise, resulting 
in wrong correspondences. 14-bit or 16-bit full radiometric data 
from a thermal infrared camera was directly utilized 
for motion estimation \cite{2019Sparse,khattak2020keyframe} to 
avoid a significant change in image appearance resulting from 
rescaling operation. However, their approaches require enabling 
NUC in long-term applications to address the temperature drift 
problem and are not directly compatible with the 8-bit image. 
By selecting the most reliable modality through several metrics, 
ROVTIO\cite{flemmen2021rovtio} fuses asynchronous thermal, visual 
and inertial measurements for estimating the odometry, which leads 
the system to autonomous switch between the different modalities 
according to the environmental conditions. 

With the development of deep learning, the neural network is 
introduced into pose estimation from thermal infrared images. 
TP-TIO\cite{zhao2020tp}, which utilizes CNN for feature detection and 
IMU-aided full radiometric-based KLT method for feature tracking, is the 
first tightly coupled deep thermal-inertial odometry algorithm. 
Combining the hallucination network with a selective fusion mechanism, 
Saputra $et$ $al$.\cite{saputra2020deeptio} proposed a deep neural odometry architecture for 
pose regression named DeepTIO, which introduced 
an end-to-end scheme. Based on DeepTIO, Saputra $et$ $al$.\cite{saputra2021graph} recently 
presented a complete thermal-inertial SLAM system, including neural 
abstraction, graph-based optimization, and a global descriptor-based 
neural loop closure detection. Combining the advantage of conventional 
and learning-based methods, Jiang $et$ $al$.\cite{jiang2022thermal} 
proposed a real-time system with an image enhancement method for 
feature detection and a light-weight optical flow network for feature 
tracking. 

The learning-based approach presents promising results. However, 
it needs GPU for algorithm acceleration and is unsuitable for 
low-cost and low-load applications. In addition, the transferability 
of network models hinders the popularization of learning-based 
methods. Considering the ground robots or MAVs in rescue applications, 
this study follows the conventional framework for real-time 
state estimation with CPU-only, and the robustness of data association 
is the core problem to be addressed.

Most existing VIO systems generally use the point features as the visual 
information. However, point features detection and tracking in 
textureless or varying illumination environments are challenging\cite{detone2018superpoint}. 
Systems integrating additional geometry structure constraints, such as line and plane information, 
were proposed to supplement the point features and improve the 
robustness of state estimation\cite{jeong2006visual,fu2020pl,zhou2019ground}.   
Edge provides semi-dense information about the environment’s 
structure and exhibits a crossover between indirect and direct 
methodologies\cite{chen2022eil}. Literature \cite{tarrio2015realtime,schenk2017robust} 
developed real-time edge-based visual odometry (REBVO) for indoor 
localization, in which edge matching is directly performed via edge contour 
alignment by minimizing distance transform (DT) error. Based on 
a local sliding window optimization over several keyframes, 
Fabian Schenk $et$ $al$.\cite{schenk2019reslam} presented the first 
edge-based SLAM system. By formulating the ICP-based motion 
estimation as maximum a posteriori (MAP) estimation, Zhou $et$ $al$.\cite{zhou2018canny} tracked edge features 
based on the approximate nearest neighbor fields.

Inspired by the significant change in thermal radiation at the edge 
of objects, our previous work developed an edge-based infrared visual 
odometry that detects and tracks the reliable edges in images 
to address the limitations of thermal infrared images in data 
association\cite{chen2022eil}. As a matter of course, this study 
presents an edge-based monocular thermal-inertial odometry (ETIO) that 
uses the edge information in the front‐end component to establish reliable 
correspondences between images. And in the back-end component, 
a pose graph optimization performs estimation over a sliding window 
of recent states by combining IMU pre-integration factors with
reprojection error of all edge feature observations. Such an optimization 
scheme can effectively suppress the influence of data interruption 
on state estimation. The main contributions of this letter are three-fold: 
\begin{enumerate}
  \item An edge-based thermal-inertial odometry, named 
  ETIO, is proposed to provide real-time state estimation 
  in visually degraded environments. To the best of our knowledge, 
  it is the first edge-based TIO and outperforms state-of-the-art 
  TIO methods.
  \item An adaptive distance transform-aided KLT (ADT-KLT) 
  tracker is proposed based on limited edge information 
  and distance field to improve the feature tracking robustness.
  \item Experiments in public datasets and the real-world show our method 
  achieves competitive accuracy and robustness in all-day state estimation. 
\end{enumerate}
\section{SYSTEM OVERVIEW}\label{section_system}
\begin{figure*}
	\centering
	\includegraphics[width=1\linewidth]{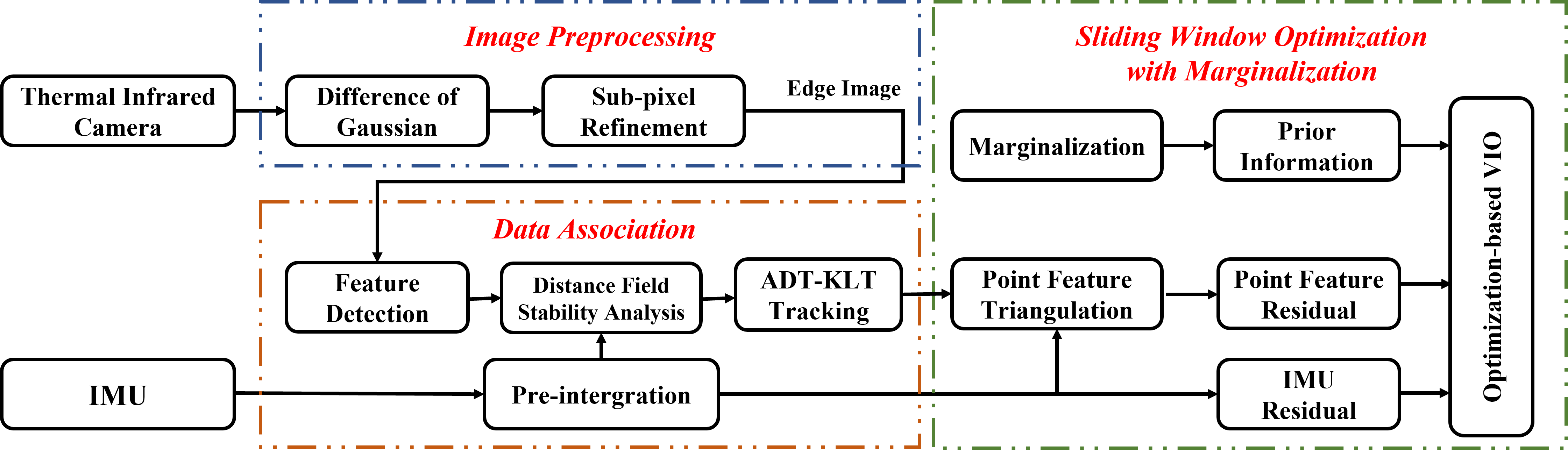}
	\caption{Overview of proposed edge-based monocular thermal-inertial odometry.}
	\label{system_overview}
  \end{figure*}

Our algorithm is specifically designed for the thermal infrared 
camera. The overall framework of the proposed ETIO consists of 
three major modules shown in \figurename{\ref{system_overview}}.
The system starts with an image preprocessing module. Thermal infrared 
images even with poor quality are converted to binarized edge images 
by the Difference of Gaussians (DoG) and sub-pixel refinement. 
As a feature enhancement algorithm, the DoG can be utilized to 
increase the visibility of edges and filter out the low-contrast 
areas. Further, sub-pixel refinement is also utilized for edge thining. 

In the data association module, the distance field of the edge image 
is utilized to introduce spatial distribution constraints and 
a stability analysis is performed to assess the consistency of 
distance field with the help of IMU pre-integration\cite{forster2015manifold}. With an adaptive 
switching policy, the ADT-KLT tracker is developed for robust 
feature tracking on edge images. Finally, a sliding-window 
optimization module is developed to tightly and efficiently 
fuse the measurement information from point features and IMU 
pre-integration. 

Frame and notation definitions used throughout this paper 
is defined as follows. We consider $(\cdot)^c$ as the camera frame, $(\cdot)^b$ 
as the body frame located at the IMU frame, and ${(\cdot)^w}$ 
as the world frame. The world frame is consistent with ${(\cdot)^b}$ at 
the initial position. $(\cdot)^{w}_{b_i}$ reflects the coordinate 
transformation from $i$th body frame to the world frame. 
To formulate the TIO optimization problem with a commonly-used 
sliding window\cite{qin2018vins}, we first define the IMU 
state at time $t_i$ as
\begin{equation}
	\begin{aligned}
	x_{i}&= [p_{b_i}^{w},q_{b_i}^{w},v_{b_i}^{w},b_{ai},b_{gi}],
	\label{state_vector}
	\end{aligned}
\end{equation}
with the position $p_{b_i}^{w}$, orientation $q_{b_i}^{w}$, 
velocity $v_{b_i}^{w}$ and biases of the accelerometer and 
gyroscope $b_{ai}$ and $b_{gi}$, respectively.  

Together with the point features parameterized by the inverse 
depth $\lambda$, the full state vector $\mathcal{X}$ to be estimated
is defined as follows
\begin{equation}
	\begin{aligned}
	\mathcal{X}=[x_{1},x_{2},\cdots,x_{n_{p}}, \lambda_{1},\lambda_{2},\cdots,\lambda_{n_{inv}}],
	\label{full_vector}
	\end{aligned}
\end{equation}
with a sliding window of $n_{p}$ states and $n_{inv}$ edge point features.
\section{METHODOLOGY}\label{methodology}
This section presents the details of ETIO shown in \figurename{\ref{system_overview}}. 
Firstly, the salient edge points for each new thermal infrared image 
frame are extracted to filter the noisy or low-contrast areas. 
Then, the existing features are tracked, and new features 
are detected to maintain a minimum number of features. An 
ADT-KLT tracker based on Distance Transform is presented 
for robust data association on sparse edge images. Finally, 
a sliding window-based tightly coupled framework is utilized 
for high accuracy and efficient state estimation. 
\subsection{Edge Extraction}
Robustness and repeatability of edge extraction in consecutive 
frames are essential to implement robust feature associations. 
The accuracy of the 2D edges in the thermal infrared image frame is also 
an important factor for state estimation accuracy. In this study, 
the DoG edge detector \cite{theoryOfEdgeDetection} is 
applied to filter noise and low contrast area given as
\begin{equation}\label{DoG}
	\begin{split}
	&\Gamma_{\sigma, k\sigma}=I*[G(x,y,\sigma)-G(x,y,k\sigma)],\\
	&s.t. \ \  G(x,y,\sigma) = \frac{1}{\sigma \sqrt{2\pi}}e^{-\frac{x^2+y^2}{2\sigma^2}},
\end{split}
\end{equation}
where the $\Gamma_{\sigma, k\sigma}$ represents the image $I$ convoluted to 
the difference of two Gaussian blur with kernel $\sigma$ and $k\sigma$, $k>1$. 
It can not only enhance edge visibility but also repress image noise. 
The gradient of the image is also calculated in the filtered image by using 
a small sigma Gaussian filter, and then the wrong edges derived from image noises and 
low contrast areas are eliminated by thresholding the gradient magnitude. 

\begin{figure}[!ht]\centering
	\includegraphics[width=8cm]{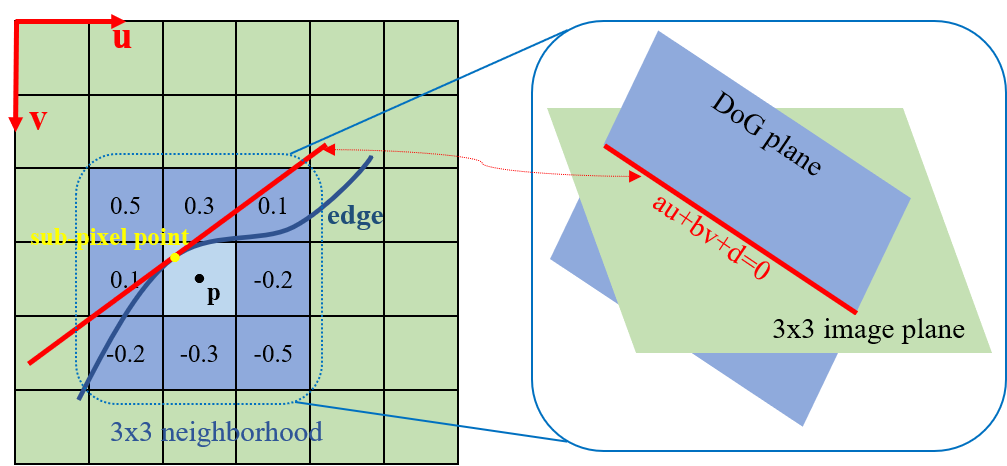}
	\caption{Sub-pixel edge extraction. The plane fitting of an edge point 
	$p$, where the red line represents the zero value line crossing 
	the DoG plane, and the yellow dot represents the sub-pixel coordinates of 
	the edge point.}\label{DoGPlaneAndDescriptor}
\end{figure}

To achieve better extraction accuracy, a 3D plane that fits the DoG 
values within a neighborhood of 3x3 pixels is expressed as 
\begin{align}\label{Sub-pixel}
	au+bv+cz+d=0
\end{align} 
where $(u,v)$ is the 2D image coordinates, and $z$ is the corresponding 
DoG value. As shown in \figurename{\ref{DoGPlaneAndDescriptor}}, the 
sub-pixel edge point is the vertical projection point of the edge's 
center on the DoG zero-crossing line. Edge gradient is obtained through 
the equation coefficients $(a,b)$. If an edge point's sub-pixel 
is located more than half a pixel away from the edge point's 
center, the edge point is eliminated. Based on the DoG-based 
edge extraction, the pixels in an edge are connected by searching 
the points within eight neighborhoods. Finally, we obtain 
the accurate position of each edge point.
\subsection{Distance Transform-aided KLT Feature Tracking}
Edge association of edge images is more challenging than 
normal images. Descriptors with enough discrimination cannot 
be extracted from the sparse binarization image, which causes 
the known descriptor-based matching\cite{campos2021orb} infeasible. The standard 
KLT tracker\cite{qin2018vins} uses spatial intensity information to establish 
point correspondences based on the brightness constancy 
assumption. But for edge images, the content of image patches is 
binarized and always sparse. Such degradation may cause the KLT 
tracker to fall into local optima, resulting in feature tracking 
errors. Therefore, this paper presents a new KLT-based tracker 
to deal with the lack of photometric information in edge 
images by introducing the Distance Transform called DT-KLT 
tracker. 

\begin{figure}[!ht]
	\centering
	\subfloat[{Distance transform generation.}]
	{
		\label{DistanceTransform}
		\includegraphics[width=8cm]{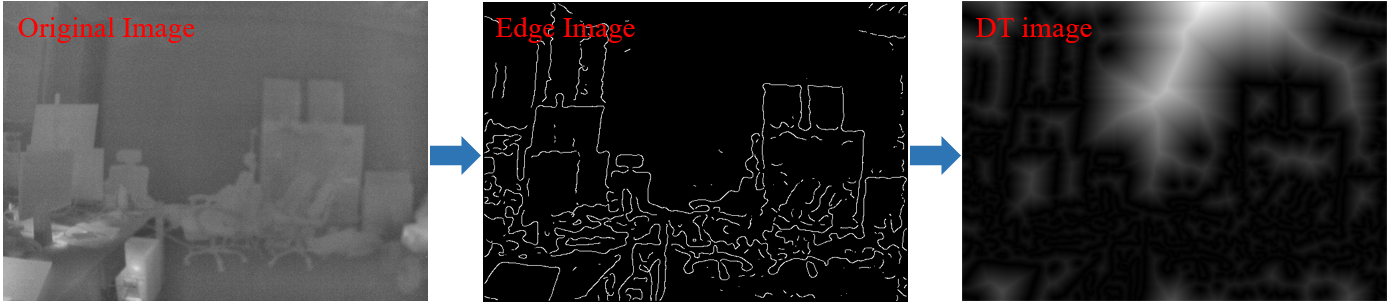}
	}
	 
	\subfloat[{pyramidal implementation.}]
	{
		\includegraphics[width=8cm]{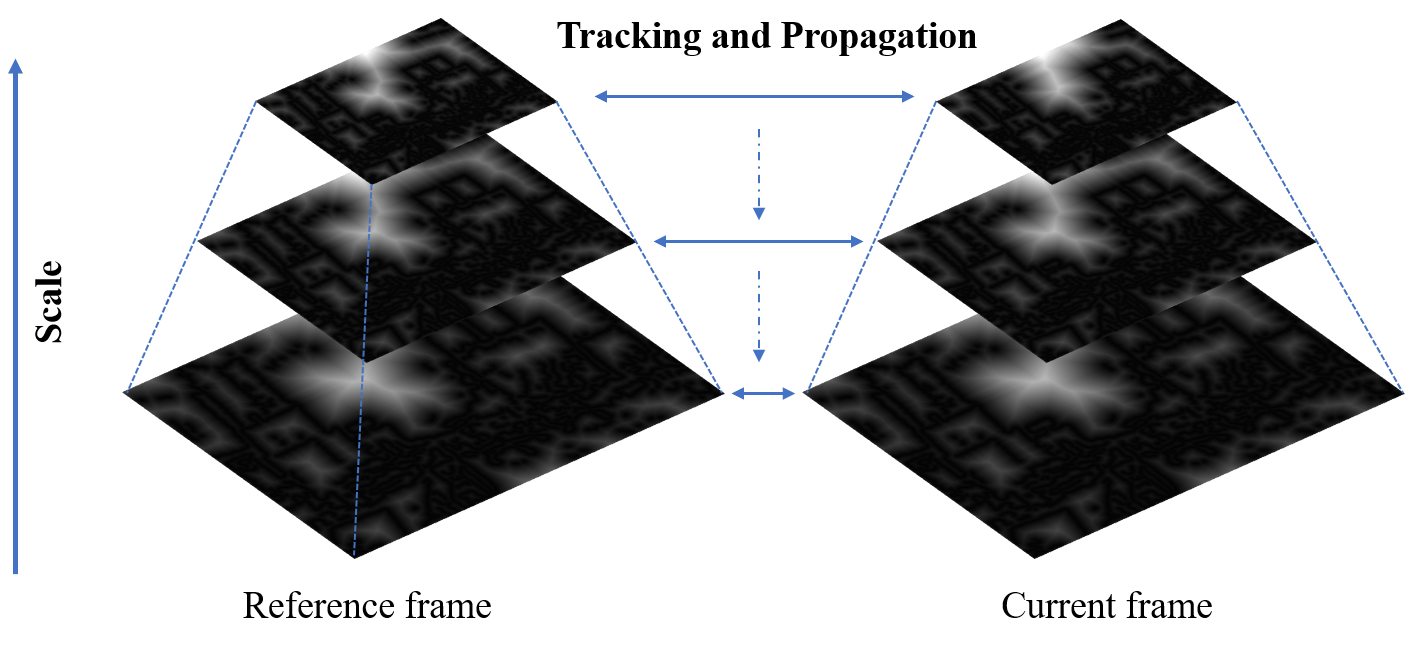}
		\label{DT-KLT}
	}
	\caption{Distance transform-aided KLT tracker.}
	\label{Vignette_calibration}
\end{figure}

Distance Transform is an operator normally applied to binary 
images, describing the two-dimensional spatial distribution of 
edge points and usually appear as a distance field. As shown 
in \figurename{\ref{DistanceTransform}}, the Distance Transform $D(p)$ 
gives edge image a dense representation, in which the gray level is defined as the 
minimal distance to the nearest edge point for every pixel $p$
\begin{align}\label{DT}
	D(p)=min\{d(p,q)|q\in I_e\},
\end{align}
where $d(p,q)$ is the Euclidean distance between $p$ and the closest edge point $q$, and $I_e$ is the binarized edge image.

Considering an edge point $u=[u_x,u_y]$ on the reference frame $I_{e}^{r}$, the goal of DT-KLT tracker is to 
find the point $v= [u_x+d_x,u_y+d_y]$ on the current frame $I_{e}^{c}$ for maximum 
similarity. The 2D position displacement $d = [d_x,d_y]$ can be solved 
by minimizing the similarity function 
\begin{align}\label{KLT_1}
\mathop{\arg\min}_{d} \sum_{x = u_x-w_x}^{u_x+w_x}\sum_{y = u_y-w_y}^{u_y+w_y}[ D_{r}(x,y)- D_{c}(x+d_x,y+d_y)]^{2},
\end{align}
where $D_r$ and $D_c$ are the corresponding distance field for 
$I_{e}^{r}$ and $I_{e}^{c}$. The above equation minimizes the 
distance distribution difference between two image patches of 
size ($2w_x+1,2w_y+1$) in distance field. A small patch size 
would be preferable for tracking accuracy, but difficult to handle 
large motions. For a trade-off between accuracy and robustness, 
image pyramidal implementation is utilized. The feature tracking is 
propagated from the highest $L_m$ to the lower level $L_{m-1}$ in scale‐space 
and so on up to the lowest $L_0$ as shown in \figurename{\ref{DT-KLT}}.
\begin{figure}[!ht]
	\centering
	\subfloat[{Standard KLT tracker applied to the original images.}]
	{
		\label{klt-gray}
		\includegraphics[width=8cm]{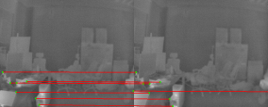}
	}
	 
	\subfloat[{Standard KLT tracker applied to the edge images.}]
	{
		\includegraphics[width=8cm]{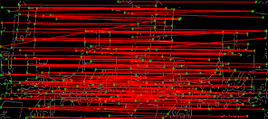}
		\label{klt-edge}
	}

	\subfloat[{DT-KLT tracker applied to the edge images.}]
	{
		\includegraphics[width=8cm]{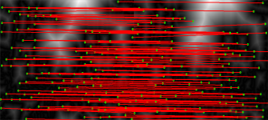}
		\label{dtklt-edge}
	}
	\caption{Feature tracking comparison with different schemes.}
	\label{KLT-result}
\end{figure}

The feature tracking performance with different schemes is shown 
in \figurename{\ref{KLT-result}. The standard KLT tracker can only 
track very few features for original thermal infrared images because of 
high image noise and low contrast. After utilizing the edge image for 
the association, tracking performance is improved significantly, but 
several wrong associations still exist due to sparse edge information. 
In comparison, our proposed DT-KLT tracker performs much better 
in tracking both tracking quantity and quality.

\subsection{Adaptive Feature Tracking Scheme}
The previous subsection proves that applying distance transform 
can benefit the feature tracking performance on edge images. 
Compared with the standard KLT tracker, the DT-KLT algorithm is 
based on the distance field constancy assumption between consecutive 
frames. However, such an assumption will be broken under the following 
cases: 1) The camera's view angle changes aggressively; 2) environmental 
photometric changes significantly. Both cases impact the edge 
extraction's consistency and significantly change the distance 
field. Taking the photometric change as an example, the appearance of 
two consecutive frames is significantly different due to the abrupt 
change in environmental illumination. As shown in \figurename{\ref{TABLE_IMAGE}, 
the discontinuous edge extraction breaks the distance field consistency 
between the reference frame and current frame, resulting in the 
degradation of DT-KLT performance. Similarly, the more aggressive 
the camera rotates, the more unstable the distance field is.
\begin{figure}[t]\centering
	\includegraphics[width=8cm]{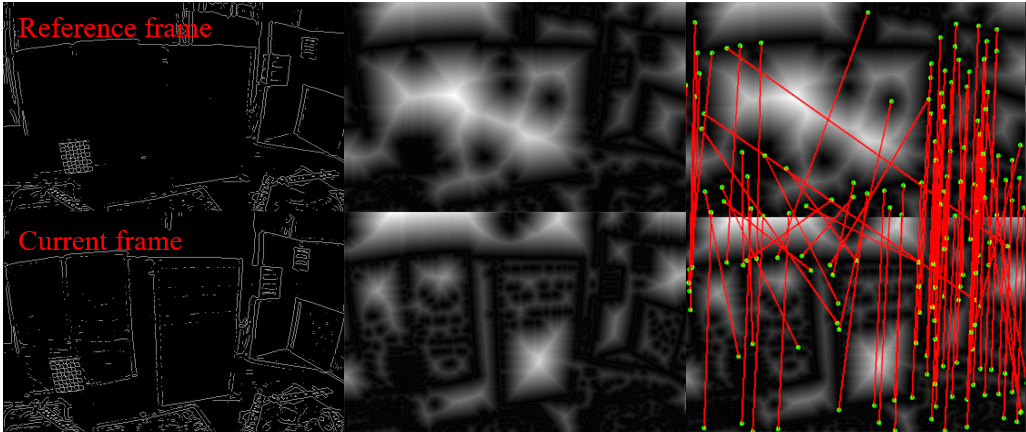}
	\caption{Feature tracking performance under discontinuous edge 
	extraction. The three columns represent the edge images, the corresponding DT images, 
	and the results of DT-KLT tracking on the distance field, respectively.}
	\label{TABLE_IMAGE}
\end{figure}

Thus, the data association scheme affects the feature tracking robustness during 
thermal infrared camera motion and subsequently influences the accuracy of the odometry. 
In response to these cases by combining the standard KLT tracker and the proposed DT-KLT 
tracker, an adaptive feature tracking pipeline ADT-KLT is developed to switch 
data association schemes according to a distance field stability 
metric $S_{df}$ designed as
\begin{equation}\label{threshold}
	S_{df}(I^c,I^r) = \alpha \cdot D_{edge}(I^c_e,I^r_e ) + \beta \cdot D_{angle}(I^c,I^r)
\end{equation}
in which $\alpha$ and $\beta$ are weight coefficients, and $D_{edge}(I^c_e,I^r_e)$ represents 
the normalized difference in the number of edge points extracted in the $I^c$ and 
$I^r$ as  
\begin{equation}\label{edgepoint}
	D_{edge}(I^c_e,I^r_e) = \frac{|num(I^c_e) - num(I^r_e)|}{num(I^r_e)}. 
\end{equation}
The $D_{angle}(I^c,I^r)$ is defined as the relative rotation 
angle between $I^c$ and $I^r$. According to Rodrigues' rotation 
formula, the relative rotation matrix $R\in SO(3)$ can be described 
as the rotation angle counterclockwise about a specified axis. 
The $D_{angle}(I^c,I^r)$ can be solved as 
\begin{equation}\label{rotationAngle}
	D_{angle}(I^c,I^r) = |arccos \frac{tr(R_r^c)-1}{2}|, 
\end{equation}
in which $tr(\cdot)$ is the trace of the matrix, and $R_r^c$ can be gained from 
IMU pre-integration module, given as
\begin{equation}\label{pre-integration}
	R_r^c = \prod_{k=1}^{n}{exp(((w_k-b_g-\eta_g)\cdot \Delta t)^{\land})}, 
\end{equation}
where $n$ is the number of IMU measurements between $I^c$ and $I^r$, 
$w_k$ is the $k$th gyroscope measurement, $b_g$ is the gyroscope bias, 
and $\eta_g$ is the Gaussian white noise, $\Delta t$ is the duration 
between two IMU measurements, and $(\cdot)^{\land}$ satisfies
\begin{equation}\label{Cross product}
	[a_1,a_2,a_3]^{\land}= \begin{bmatrix}
		   	      0  & -a_3&a_2  \\
		   	    a_3  & 0  &-a_1 \\
		   	    -a_2 & a_1&0		
		   	    \end{bmatrix}.
\end{equation}

More specifically, a larger value of $S_{df}$ means that the dynamic change 
of the current distance field is more significant. Hence the adaptive switching 
policy is that when $S_{df}<S_{th}$, the DT-KLT tracker is utilized on edge images 
association. Otherwise, a standard KLT tracker is utilized instead. $S_{th}$ is the adaptive 
threshold. The ablation experiment in the Subsection \ref{dataset} illustrates this adaptive policy's 
effects, which helps improve localization accuracy. 
\subsection{Tightly-Coupled Formulation}\label{Tight-couple}
Robust edge feature tracking between thermal infrared frames 
is completed with the proposed ADT-KLT tracker, which is 
further utilized for reprojection error as 
\begin{align}\label{Visual Residual}
	\boldsymbol {r}_{C}(\mathcal{X})=[b_1\ b_2]^T \cdot (\hat{\bar{\mathcal{P}}}_{l}^{c_j}-\frac{\mathcal{P}_{l}^{c_j}} {\| \mathcal{P}_{l}^{c_j} \|}),
\end{align}
where $\mathcal{X}$ is the variables to be estimated defined in Eq.\eqref{full_vector}, and $\mathcal{P}_{l}^{c_j}$ and $\hat{\bar{\mathcal{P}}}_{l}^{c_j}$ 
are the projected vector on the unit sphere of $p_{l}^{c_i}$ and $p_{l}^{c_j}$, 
respectively; $p_{l}^{c_i}$ and $p_{l}^{c_j}$ are the $l$th edge point tracked 
in the $i$th and $j$th frames by ADT-KLT tracker; $b_1$ 
and $b_2$ are two orthogonal bases of any choice that span the 
tangent plane of $\hat{\bar{\mathcal{P}}}_{l}^{c_j}$.

With the developed robust edge association, a bundle adjustment 
strategy is then utilized to fuse the measurements from IMU 
and thermal infrared images and implement the estimation of 
state $\mathcal{X}$. 
 
The residuals for visual and IMU measurement, ${r}_{C}(\mathcal{X})$ 
and ${r}_{B}(\mathcal{X})$, and the prior information from 
marginalization $r_{p}(\mathcal{X})$ are minimized by sliding 
window optimization based on MAP. The objective function for the joint optimization is 
designed as follows
\begin{equation}\label{Localization}
	\begin{aligned}
\mathop{\arg\min}_{\mathcal{X}} \bigg \{\|r_{p}(\mathcal{X})\|^2+\sum_{}\|{r}_{\mathcal{B}}(\mathcal{X})\|^2_{\scriptscriptstyle{\sum_{b}}}+ \sum_{}\rho(\|{r}_{\mathcal{C}}(\mathcal{X})\|^2_{\scriptscriptstyle{\sum_{c}}}) \bigg\},
	\end{aligned}
\end{equation}
where $\sum_{b}$ and $\sum_{c}$ represent the measurement 
covariance matrices of the IMU and edge point, respectively.
In addition, $\rho(\cdot)$ is the Huber norm function. And 
then, the Ceres solver\cite{Agarwal_Ceres_Solver_2022} is 
used to solve this nonlinear optimization problem in 
Eq.\eqref{Localization}. For more details about ${r}_{B}(\mathcal{X})$ 
and $r_{p}(\mathcal{X})$, please refer to \cite{qin2018vins}.
\section{EXPERIMENTS}\label{experiment}
The proposed approach was verified with several experiments on public datasets 
and in the real world. As shown in \figurename\ref{robot}, our self-designed 
tracked robot was utilized for real-world experiments with a multi-sensors device 
consisting of a GNSS device, a 3D Lidar, an IMU, a visual camera, and a thermal infrared camera. 
The corresponding specifications are listed in Table \ref{table_sensor}.
In addition, the onboard computer is an Intel NUC with an Intel i7-10710U 
processor.
\begin{figure}[!ht]\centering
	\includegraphics[width=7cm]{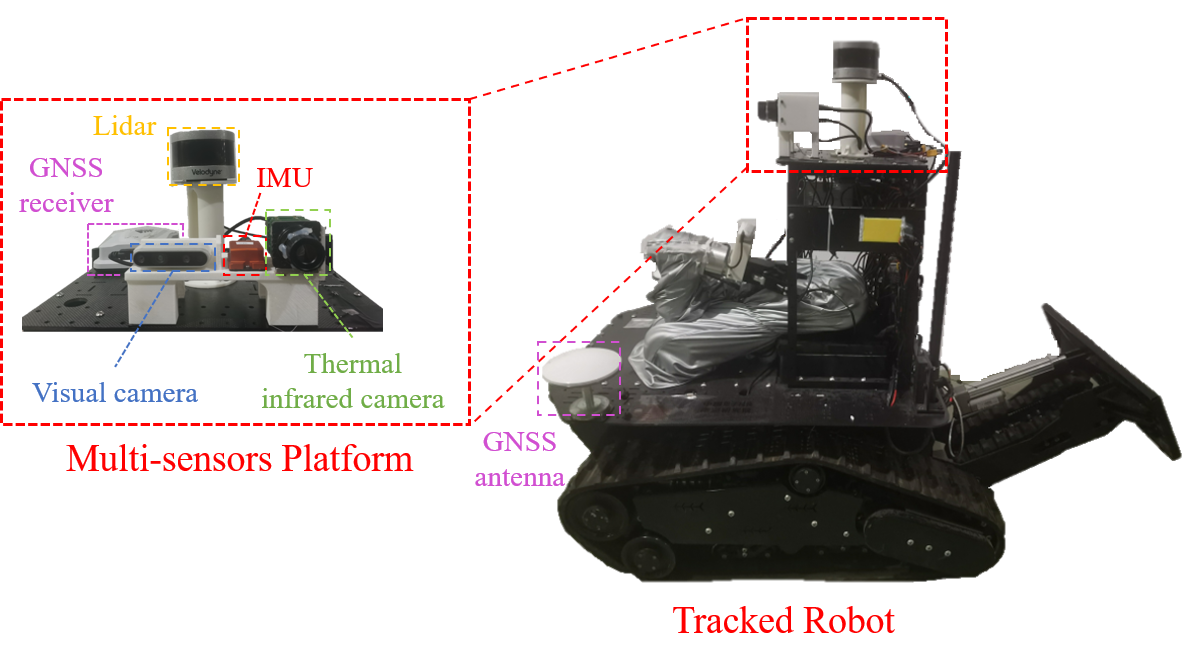}
	\caption{Experimental platform for outdoor localization.}
	\label{robot}
\end{figure}
\begin{table}[htbp]\centering	
	\scriptsize
	\setlength\tabcolsep{2pt}
	\centering	
		\caption{Specifications of the sensors.}
		\label{table_sensor}       
		\begin{tabular}{ccccc}
			\hline\noalign{\smallskip}
			Type   & Model & Description  \\ 
			\noalign{\smallskip}\hline\noalign{\smallskip}
			Lidar   &Velodyne VLP-16 & $360^\circ$ FOV with 10Hz  \\
			IMU     &MTi-G-710 & 200Hz\\
			GNSS     &NovAtel  & 20Hz\\
			Visual Camera &Intel D435i & 640$\times$480px with 25Hz  \\
			Thermal Infrared Camera &Gobi+ 640  & 640$\times$480px with 50Hz\\
			\noalign{\smallskip}\hline
		\end{tabular}
\end{table}
\subsection{Evaluation on the Public Datasets}\label{dataset}
We first evaluated our proposed approach on the Urban Parking Lot 
Dataset and the Active Gold Mine Dataset provided in 
\cite{khattak2020keyframe}. Both datasets are collected by 
a DJI Matrice M100 quadrotor platform and combine thermal 
camera (Tau2), IMU (VN-100), and LiDAR (OS-1) measurements.

We compared our approach with state-of-the-art solutions, 
i.e. VINS-Mono\cite{qin2018vins}, PL-VINS\cite{fu2020pl}, 
ROTIO\cite{khattak2020keyframe,flemmen2021rovtio}, and ORB-SLAM3\cite{campos2021orb}. 
ROTIO is a modified version of ROVIO\cite{bloesch2017iterated} 
for thermal infrared camera applications utilizing full 
radiometric data for initialization and tracking. The parameters 
of compared algorithms were set to the default values in the open-source 
codes and all results were obtained without loop-closure except for ORB-SLAM3. 
The latest LiDAR-inertial odometry FAST-LIO2\cite{xu2022fast} was utilized as the ground truth. 
The accuracy was evaluated by Root-Mean-Square-Error (RMSE) of Absolute Trajectory 
Error (ATE).
\renewcommand\arraystretch{1.2}
\newcommand{\tabincell}[2]{\begin{tabular}{@{}#1@{}}#2\end{tabular}}  
\begin{table*}[!ht]
	\centering
	\caption{RMSE ATE [m] on the public Datasets\cite{khattak2020keyframe}. Best results are in bold. The Blank (-) represents failure.}
	\label{table_TIO}
	\begin{tabular}{cccccccc}
	\hline\noalign{\smallskip}
	\multirow{2}{*}{Dataset} &\multirow{2}{*}{ORB-SLAM3}&\multirow{2}{*}{VINS-Mono}&\multirow{2}{*}{PL-VINS} & \multirow{2}{*}{ROTIO}&\multicolumn{3}{c}{ETIO}\\
	\cline{6-8}
	
    & & & & &KLT & DT-KLT & ADT-KLT\\
	\hline\noalign{\smallskip}
	the Urban Parking Lot &- & 4.892 & 3.594	&2.934	&2.136	&1.049&$\mathbf{0.648}$\\
	the Active Gold Mine &-  & 5.982 & 1.895	&0.993	&1.191	&0.472&$\mathbf{0.409}$\\
	\hline\noalign{\smallskip}
\end{tabular}
	\end{table*}

Figure \ref{Charlie} presents the comparative trajectories. Table \ref{table_TIO} summarizes 
the estimation errors. The results show that our method outperforms 
the other open-source solutions. Because of low contrast and high 
noise, ORB-SLAM3 failed to provide completed trajectory 
information in both datasets. It is seen that the lack of 
distinctive feature detection increases the number of false positives in the 
descriptor matching. VINS-Mono relying on optical flow for 
feature tracking, performed robust to image quality but is 
susceptible to photometric changes suddenly caused by NUC 
or rescaling operation. PL-VINS exploited structural constraints for a 
supplement but with limited performance enhancement. The 
proposed ETIO performed the best in both datasets, benefiting from 
the robust edge-based data association scheme. Inferring that the impact of 
noise and NUC is relatively small for the full radiometric 
data but not negligible, ROTIO performed better than conventional 
VIO methods but worse than ETIO.
\begin{figure}[!ht]
	\centering
	\subfloat[{comparison the Urban Parking Lot dataset.}]
	{
		\label{parking_lot}
		\includegraphics[width=7cm]{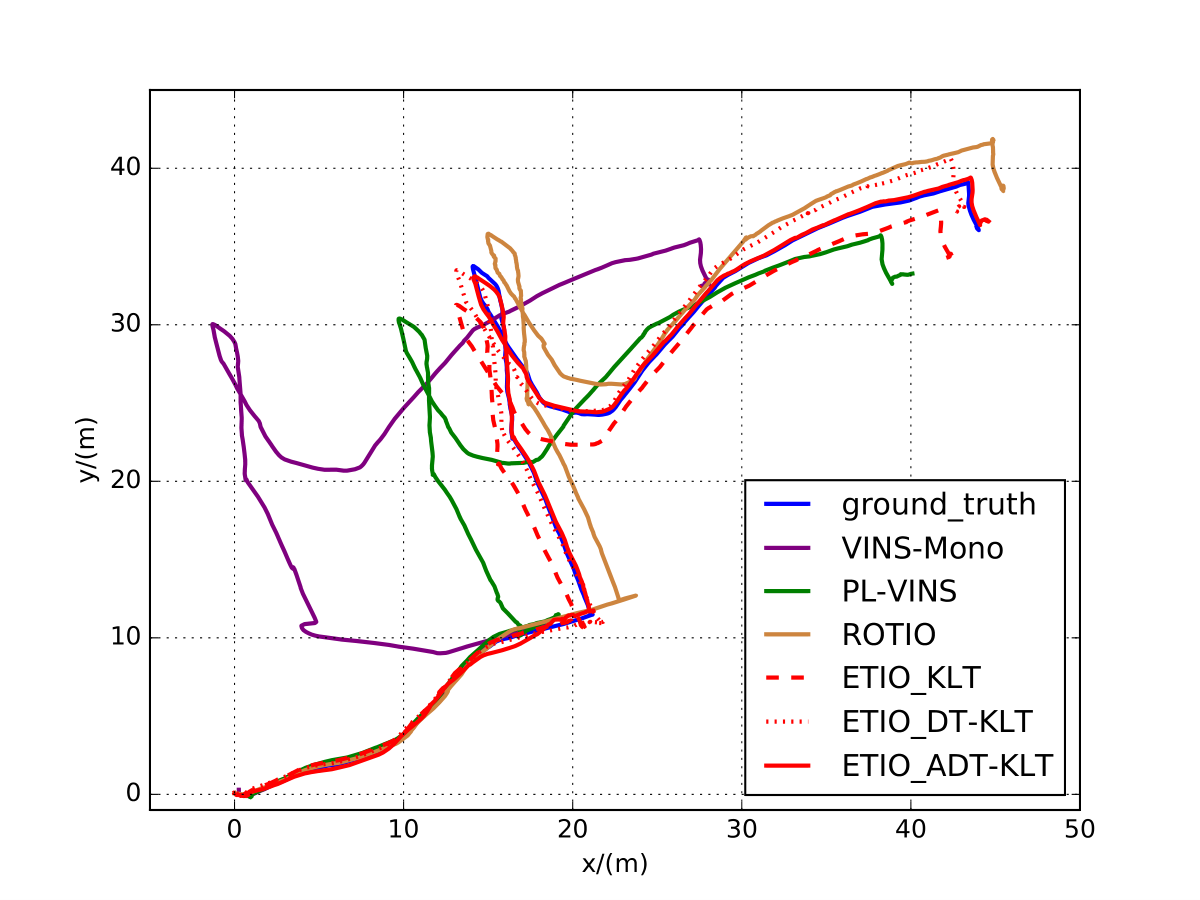}
	}
	 
	\subfloat[{comparison the Active Gold Mine dataset.}]
	{
		\includegraphics[width=7cm]{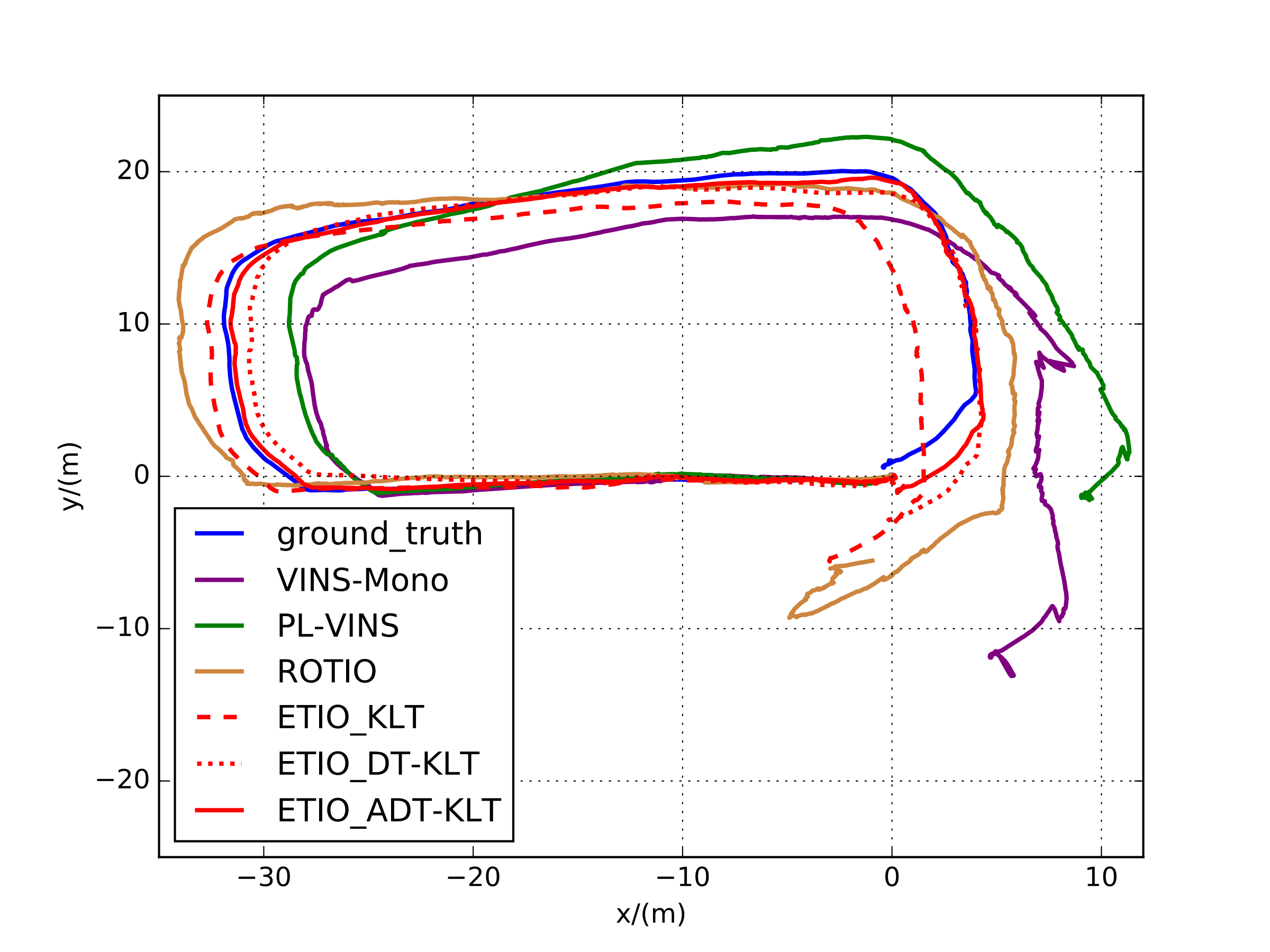}
		\label{mine}
	}
	\caption{Trajectory comparisons on the public dataset.}
	\label{Charlie}
\end{figure}

Then, we performed an ablation study to show the effects of the 
proposed ADT-KLT tracker. We evaluated the performance of the 
three data association methods, i.e., the standard KLT tracker, 
the DT-KLT tracker, and the ADT-KLT tracker. As shown in Table \ref{table_TIO}, 
both the DT-KLT tracker and ADT-KLT tracker performed better 
than the standard KLT tracker, which validated that the distance 
field can efficiently improve the edge feature tracking. ADT-KLT 
tracker obtained the best performance because the proposed adaptive 
policy can suppress the tracking loss caused by sudden changes in 
the distance field. 

To quantitatively evaluate the real-time performance of the system, 
we also recorded the time consumption of all the sequences for analysis. 
The time consumption statistics with a mean and a standard deviation 
per round for each component are shown in \figurename\ref{time}. 
It is seen that the proposed ETIO exhibits real-time performance 
because parallel computation and multithreading technologies 
are adopted for different modules. The state update frequency is 
determined by the highest execution time in the three threads. 
The maximum output frequency of the odometer is about 25Hz so that our 
approach can work in real-time without GPU acceleration.
\begin{figure}[!ht]\centering
	\includegraphics[width=7cm]{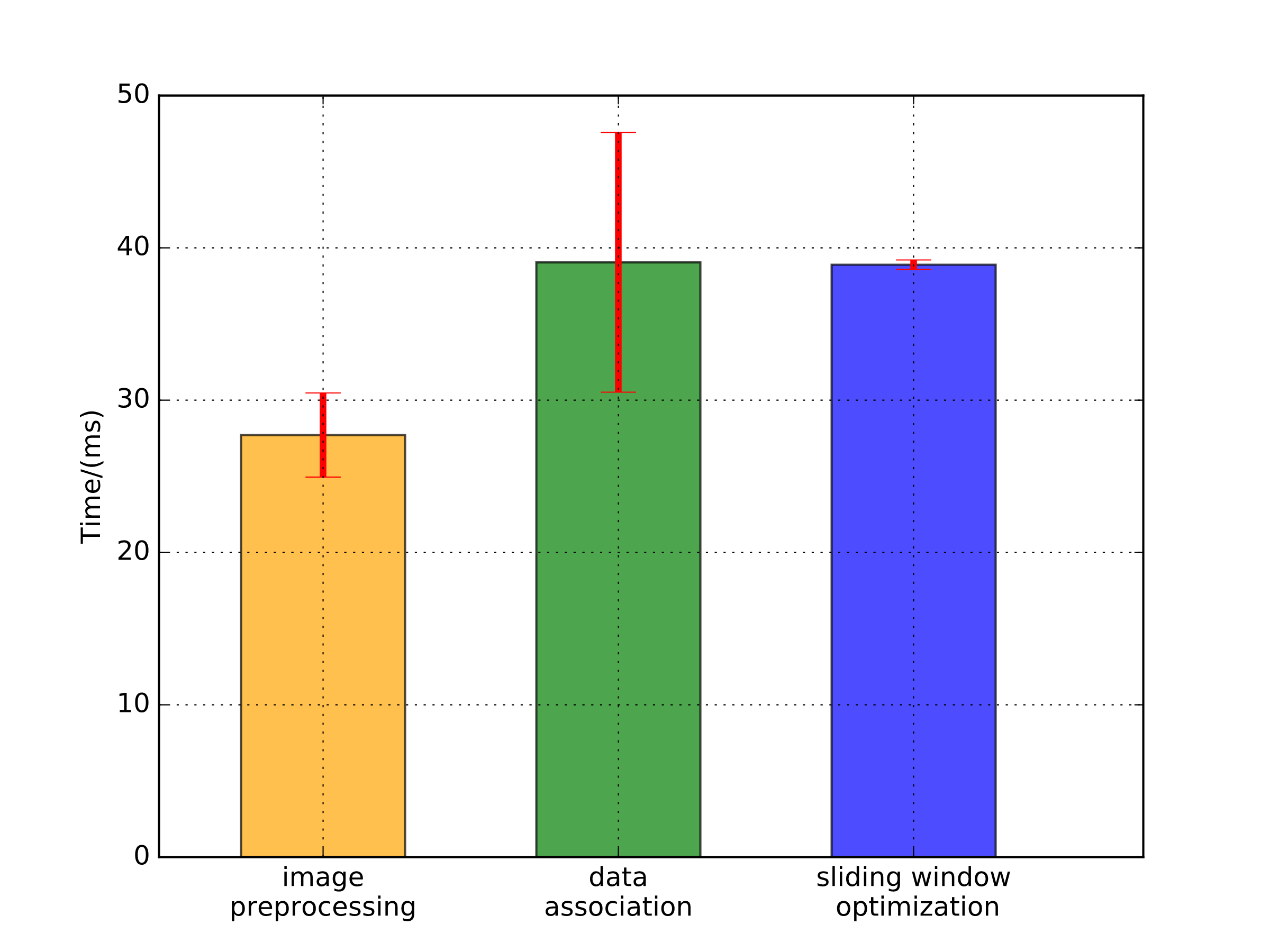}
	\caption{Time consumption statistics with mean and 1$\sigma$ bound of each step in ETIO.}\label{time}
\end{figure}

\subsection{Evaluation in Outdoor Environments}
A real-world experiment was further performed in an outdoor urban 
environment to evaluate the all-day localization accuracy of 
ETIO. The tracked robot in \figurename\ref{robot} was manually 
controlled to run similar trajectories in the park, and four sequences 
were recorded at different times of the day shown in Table \ref{table_outdoor}. 
The total path length of each sequence is approximately 200 m, and the GNSS and Lidar 
provided the ground truth of trajectories.

Unlike the scenes in the previous subsection, which are 
mainly indoors or in caves, TIO is more challenging in outdoor 
scenes where the thermal infrared image is affected by the solar 
radiation. The estimated trajectories are shown in \figurename\ref{figure_9}. 
Noticed that ROTIO performed much worse than the evaluation of 
Subsection \ref{dataset}, and even failed in Seq.01. Quantitative results of trajectory error are 
presented in Table \ref{table_outdoor}. The heat accumulation 
from solar radiation caused the image to saturate, 
resulting in inconsistent patch tracking and state estimate 
divergence. At night sequence, the sun's radiation got weaker, making 
the thermal radiation of different objects outdoors more 
distinguishable. Hence, the performance of ROTIO improved as the night fell. 
\begin{figure}[!htbp]
	\centering
	\subfloat[Trajectories of Seq.01.]{\includegraphics[width=4.25cm]{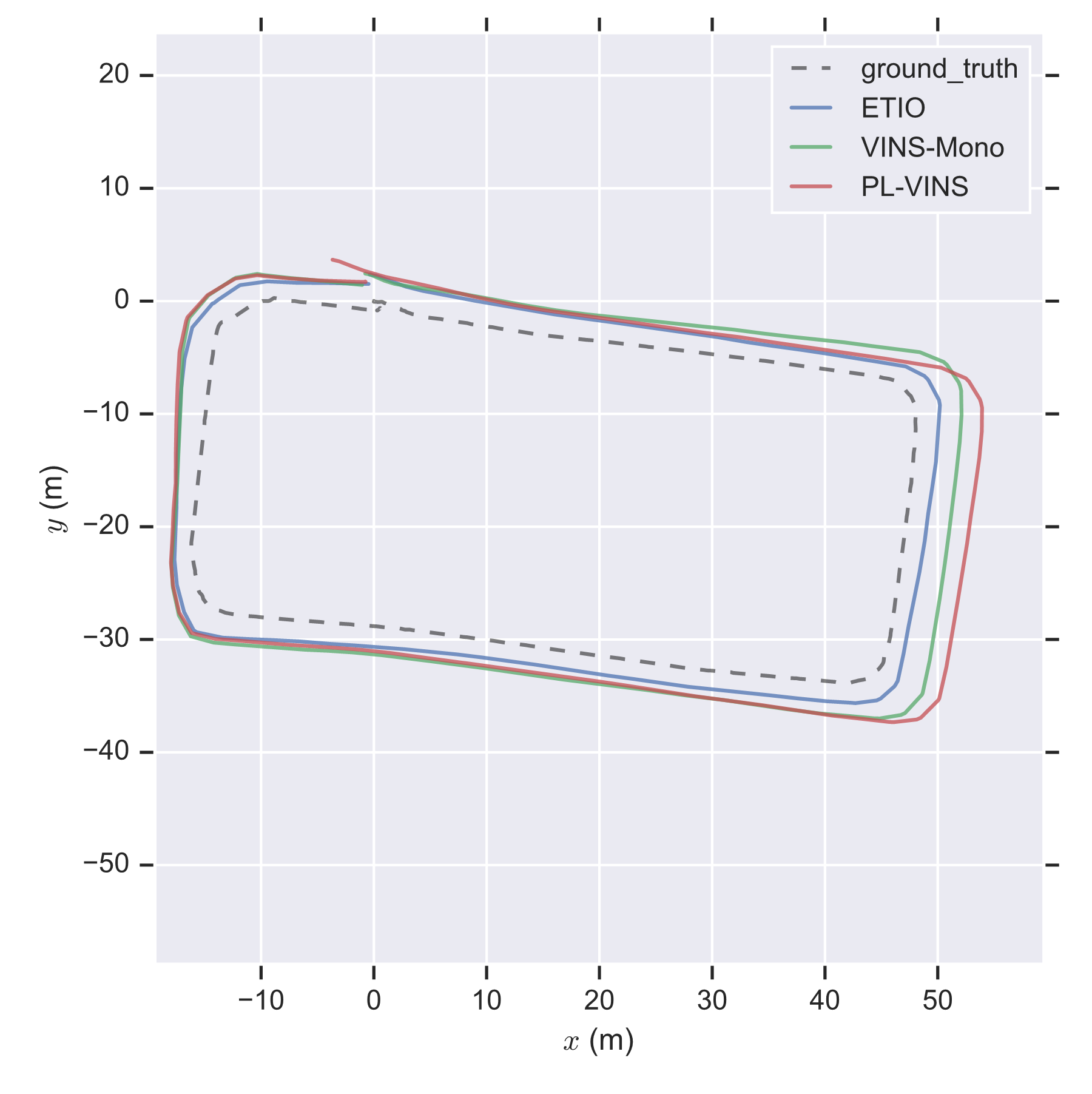}}
	\subfloat[Trajectories of Seq.02.]{\includegraphics[width=4.25cm]{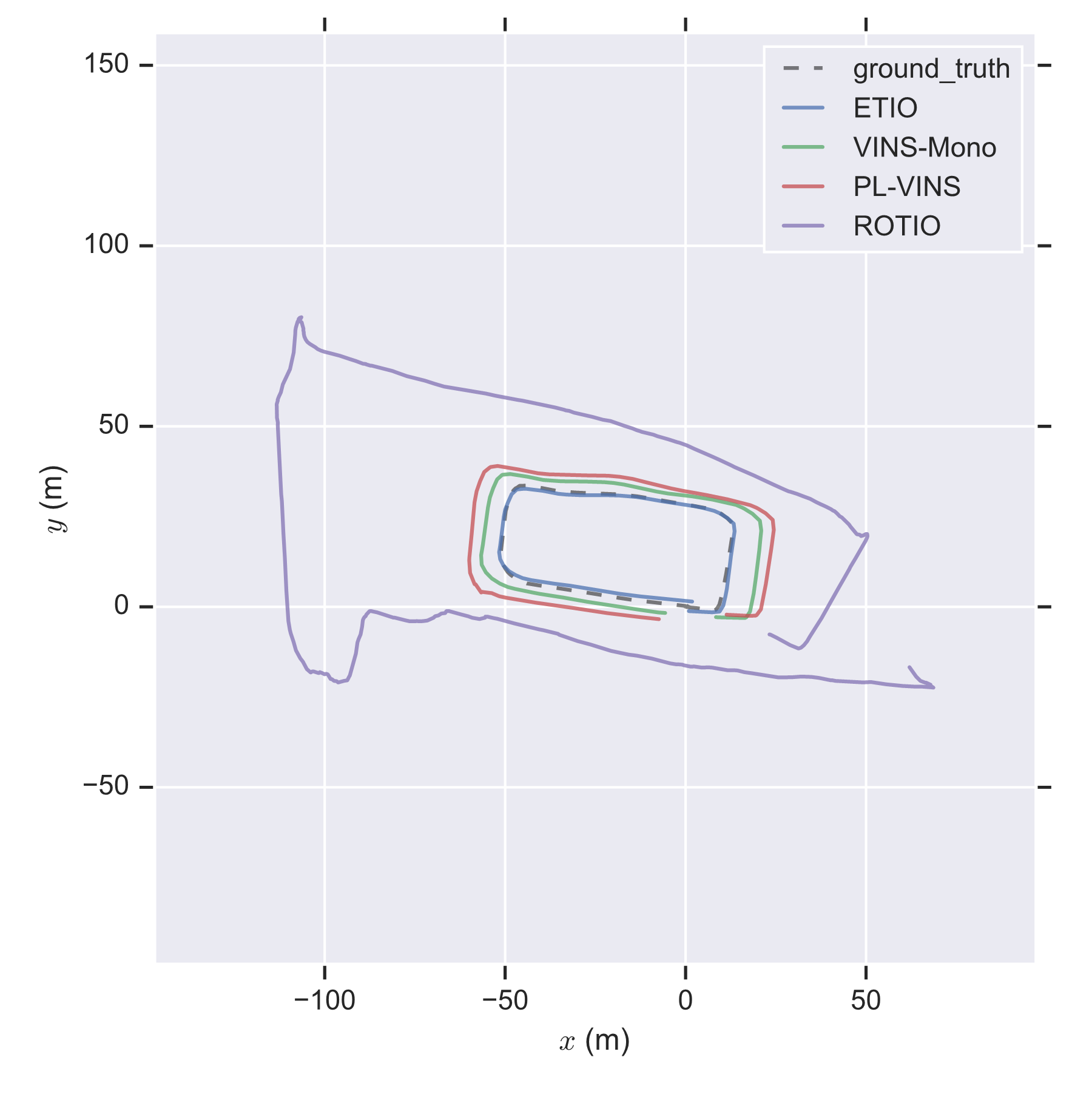}}\\
	\subfloat[Trajectories of Seq.03.]{\includegraphics[width=4.25cm]{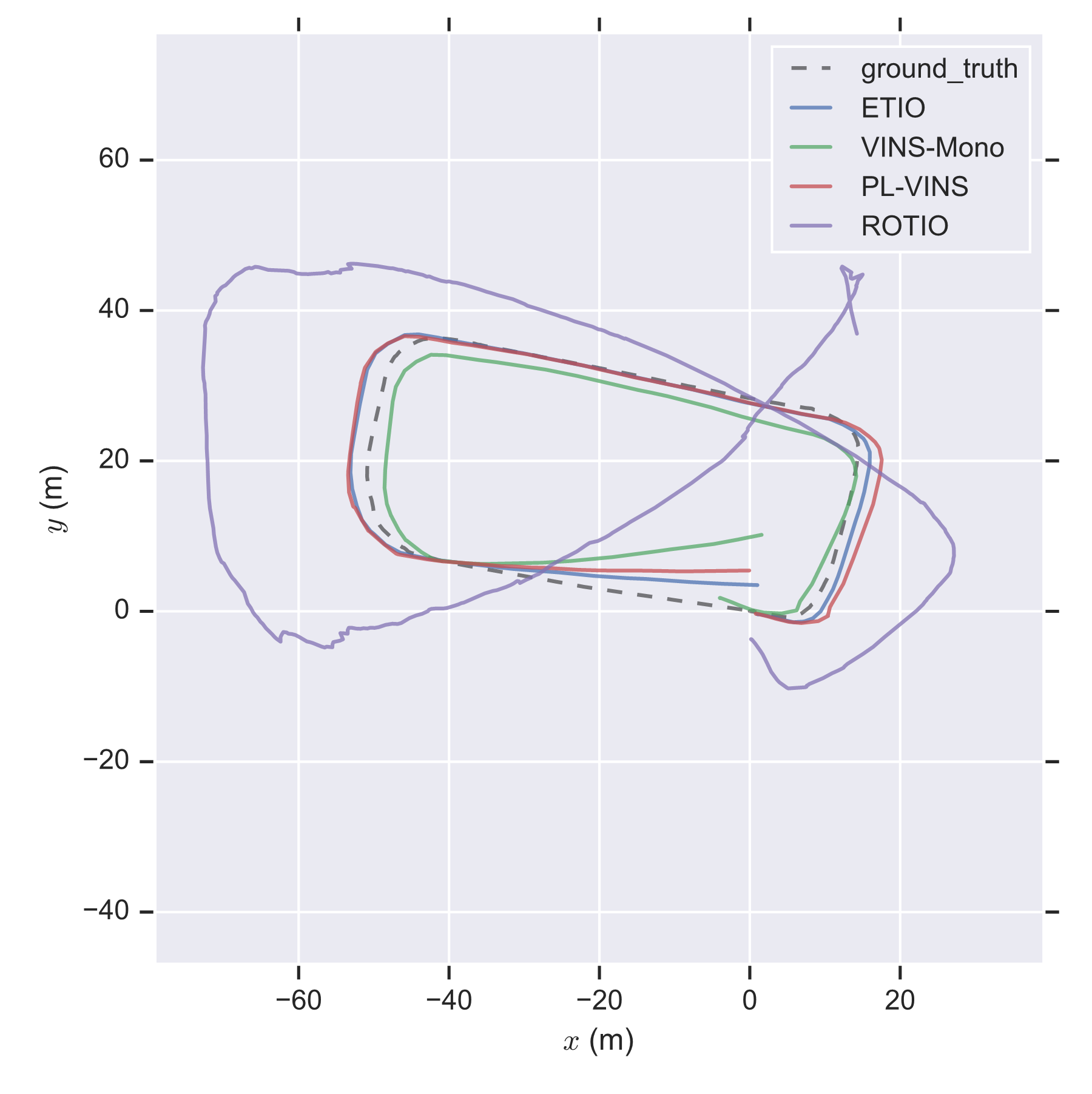}}
	\subfloat[Trajectories of Seq.04.]{\includegraphics[width=4.25cm]{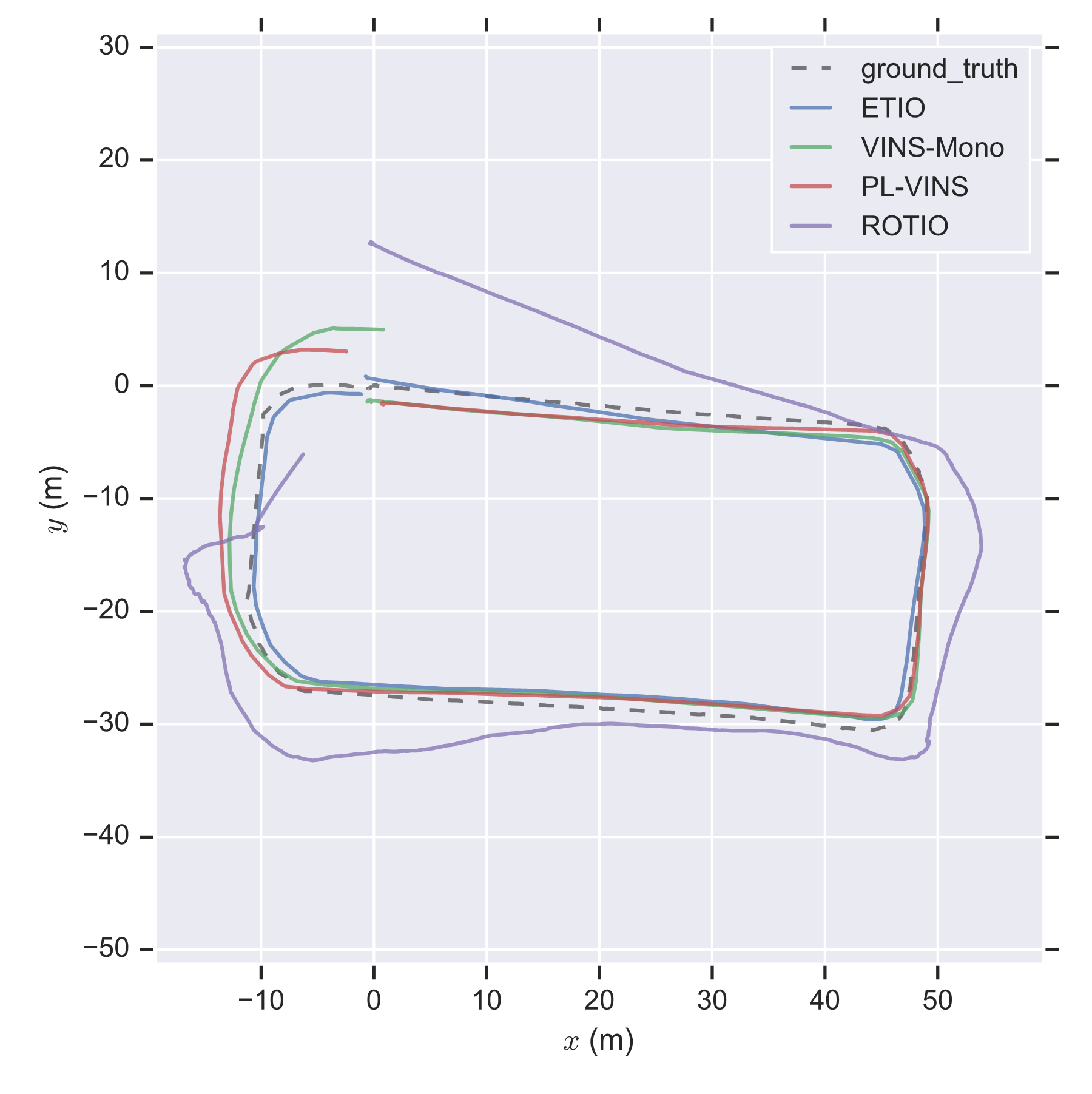}}
	\caption{Comparison of the estimated trajectories in outdoor localization experiments.}
	\label{figure_9}
\end{figure}

Compared with ROTIO, the other mentioned approaches performed 
more robust to illumination changes during the all-day time and 
provided continuous trajectories. Line feature detection and 
matching introduce errors due to low image quality. 
For this reason, PL-VINS, which makes additional use of the 
line feature constraints compared to VINS-Mono, was not performing 
better than VINS-Mono. ETIO presented the best performance 
among all the sequences, which validates that the proposed 
edge-based method achieves robust all-day localization ability.
\begin{table}[htbp]\centering	
		\small
		\setlength\tabcolsep{3pt}
		\centering	
			\caption{RMSE ATE [m] of the outdoor localization. The best results are in bold. The blank (-) represents failure.}
			\label{table_outdoor}       
			\begin{tabular}{cccccc}
				\hline\noalign{\smallskip}
				Seq&Time of day&VINS-Mono&PL-VINS&ROTIO&ETIO\\
				\noalign{\smallskip}\hline\noalign{\smallskip}
	01& Day(14:57)  & 3.301 & 4.101	&-   	&$\mathbf{2.170}$\\
	02& Dusk(18:27) & 6.263 & 9.338	&41.330	&$\mathbf{1.224}$\\
	03& Night(21:08)& 4.296 & 2.636	&18.058	&$\mathbf{1.966}$\\
	04& Night(23:15)& 2.166 & 2.286	&8.880	&$\mathbf{1.403}$\\
				\noalign{\smallskip}\hline
			\end{tabular}
	\end{table}

\subsection{Evaluation in Extreme Environment}
The proposed approach was deployed in a dark, smoke-filled room for 
evaluation in extreme illumination environments such as conflagration areas. 
The artificial smoke was added to make the environment more degraded, 
created by a theatrical smoke generator. We held the sensor platform 
in hand, walked around the specified trajectory, and eventually 
returned to the starting point. Two sequences with different path 
lengths were recorded. As VICON or Lidar-based state estimation 
approaches cannot work for ground truth output in such an environment, 
trajectory accuracy was evaluated by accumulated drift.
\begin{table}[!htbp]\centering	
	\small
	\setlength\tabcolsep{3pt}
	\centering	
		\caption{Accumulated drift [m] of localization in a drak and smoke-filled room. 
		Best results are in bold. The Blank (-) represents failure.}
		\label{tab_extreme}       
		\begin{tabular}{cccccc}
			\hline\noalign{\smallskip}
			Seq &Length (m) &VINS-Mono &PL-VINS& ROTIO &ETIO \\ 
			\noalign{\smallskip}\hline\noalign{\smallskip}
			01  &73.2 &- &-   & 1.035 &$\mathbf{0.211}$\\
			02  &54.3 &- &-   & 0.746 &$\mathbf{0.080}$ \\
			\noalign{\smallskip}\hline
		\end{tabular}
\end{table}
The error statistics are shown in Table \ref{tab_extreme}, and our method 
outperforms the others in both sequences, where VINS-Mono and PL-VINS both 
failed. Take Seq.01 as an example, and the estimated trajectories 
are shown in \figurename{\ref{Extreme}}. ROTIO achieved continuous 
estimation of camera trajectories but exhibited a significant error in the yaw 
angle estimation, especially during turning; we infer that it's 
because the thermal infrared images had a significant appearance 
change when turning, leading to data association errors. In 
contrast, ETIO performed better trajectory consistency benefitting 
from our robust edge-based data association solution.
\begin{figure}[!ht]\centering
	\includegraphics[width=7cm]{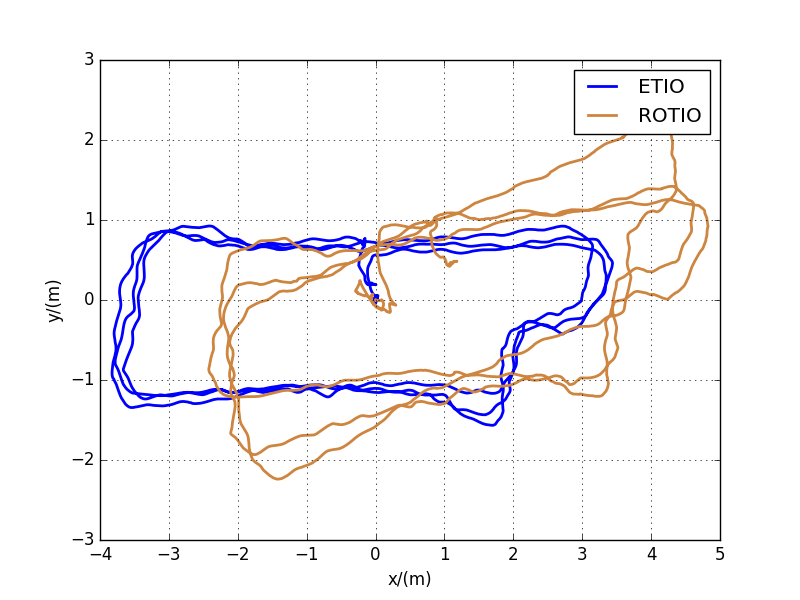}
	\caption{Estimated trajectories of Seq.01 under ROTIO and ETIO.}\label{Extreme}
\end{figure}
\section{Conclusions}
\label{section_conclusions}
Compared with the visual camera, the thermal infrared camera has 
the potential to work all-day time. Motivated by the phenomenon 
that thermal radiation varies most significantly at the 
edges of objects, the study proposes an edge-based monocular 
thermal-inertial odometry, called ETIO, to salient edge information 
for robust state estimation in visually degraded environments. 
Distance field-aided feature tracking and an adaptive switching 
policy are proposed to overcome the difficulties of sparse edge 
image data association. Extensive evaluations indicated the 
reliable and robust performance of the proposed method for state 
estimation in visually degraded environments.

Considering that the edge information is also relatively 
robust for visual images, we believe that edge features 
are the bridge of tightly coupled visual and thermal infrared 
streams for state estimation. In future work, we aim to 
exploring the applications in edge-based visual-thermal-inertial 
fusion framework.

\bibliographystyle{IEEEtran} 
\balance
\bibliography{mybib}

\begin{thebibliography}{10}
\providecommand{\url}[1]{#1}
\csname url@rmstyle\endcsname
\providecommand{\newblock}{\relax}
\providecommand{\bibinfo}[2]{#2}
\providecommand\BIBentrySTDinterwordspacing{\spaceskip=0pt\relax}
\providecommand\BIBentryALTinterwordstretchfactor{4}
\providecommand\BIBentryALTinterwordspacing{\spaceskip=\fontdimen2\font plus
\BIBentryALTinterwordstretchfactor\fontdimen3\font minus
  \fontdimen4\font\relax}
\providecommand\BIBforeignlanguage[2]{{%
\expandafter\ifx\csname l@#1\endcsname\relax
\typeout{** WARNING: IEEEtran.bst: No hyphenation pattern has been}%
\typeout{** loaded for the language `#1'. Using the pattern for}%
\typeout{** the default language instead.}%
\else
\language=\csname l@#1\endcsname
\fi
#2}}

\bibitem{qin2018vins}
T.~Qin, P.~Li, and S.~Shen, ``Vins-mono: A robust and versatile monocular
  visual-inertial state estimator,'' \emph{IEEE Transactions on Robotics},
  vol.~34, no.~4, pp. 1004--1020, 2018.

\bibitem{von2022dm}
L.~von Stumberg and D.~Cremers, ``Dm-vio: Delayed marginalization
  visual-inertial odometry,'' \emph{IEEE Robotics and Automation Letters},
  vol.~7, no.~2, pp. 1408--1415, 2022.

\bibitem{khattak2020keyframe}
S.~Khattak, C.~Papachristos, and K.~Alexis, ``Keyframe-based thermal--inertial
  odometry,'' \emph{Journal of Field Robotics}, vol.~37, no.~4, pp. 552--579,
  2020.

\bibitem{banuls2020object}
A.~Banuls, A.~Mandow, R.~V{\'a}zquez-Mart{\'\i}n, J.~Morales, and
  A.~Garc{\'\i}a-Cerezo, ``Object detection from thermal infrared and visible
  light cameras in search and rescue scenes,'' in \emph{2020 IEEE International
  Symposium on Safety, Security, and Rescue Robotics (SSRR)}.\hskip 1em plus
  0.5em minus 0.4em\relax IEEE, 2020, pp. 380--386.

\bibitem{shin2020dvl}
Y.-S. Shin, Y.~S. Park, and A.~Kim, ``Dvl-slam: Sparse depth enhanced direct
  visual-lidar slam,'' \emph{Autonomous Robots}, vol.~44, no.~2, pp. 115--130,
  2020.

\bibitem{2004Nonuniformity}
O.~Riou, S.~Berrebi, and P.~Bremond, ``Nonuniformity correction and thermal
  drift compensation of thermal infrared camera,'' \emph{Proceedings of SPIE -
  The International Society for Optical Engineering}, vol. 5405, 2004.

\bibitem{Beauvisage2016Multi}
A.~Beauvisage, N.~Aouf, and H.~Courtois, ``Multi-spectral visual odometry for
  unmanned air vehicles,'' in \emph{2016 IEEE International Conference on
  Systems, Man, and Cybernetics (SMC)}.\hskip 1em plus 0.5em minus 0.4em\relax
  IEEE, 2016, pp. 001\,994--001\,999.

\bibitem{Poujol2016A}
J.~Poujol, C.~A. Aguilera, E.~Danos, B.~X. Vintimilla, R.~Toledo, and A.~D.
  Sappa, ``A visible-thermal fusion based monocular visual odometry,'' in
  \emph{Robot 2015: Second Iberian Robotics Conference}.\hskip 1em plus 0.5em
  minus 0.4em\relax Springer, 2016, pp. 517--528.

\bibitem{article}
T.~Mouats, N.~Aouf, L.~Chermak, and M.~A. Richardson, ``Thermal stereo odometry
  for uavs,'' \emph{IEEE Sensors Journal}, vol.~15, no.~11, pp. 6335--6347,
  2015.

\bibitem{2019Sparse}
Y.-S. Shin and A.~Kim, ``Sparse depth enhanced direct thermal-infrared slam
  beyond the visible spectrum,'' \emph{IEEE Robotics and Automation Letters},
  vol.~4, no.~3, pp. 2918--2925, 2019.

\bibitem{flemmen2021rovtio}
H.~D. Flemmen, ``Rovtio: Robust visual thermal inertial odometry,'' Master's
  thesis, NTNU, 2021.

\bibitem{zhao2020tp}
S.~Zhao, P.~Wang, H.~Zhang, Z.~Fang, and S.~Scherer, ``Tp-tio: A robust
  thermal-inertial odometry with deep thermalpoint,'' in \emph{2020 IEEE/RSJ
  International Conference on Intelligent Robots and Systems (IROS)}.\hskip 1em
  plus 0.5em minus 0.4em\relax IEEE, 2020, pp. 4505--4512.

\bibitem{saputra2020deeptio}
M.~R.~U. Saputra, P.~P. de~Gusmao, C.~X. Lu, Y.~Almalioglu, S.~Rosa, C.~Chen,
  J.~Wahlstr{\"o}m, W.~Wang, A.~Markham, and N.~Trigoni, ``Deeptio: A deep
  thermal-inertial odometry with visual hallucination,'' \emph{IEEE Robotics
  and Automation Letters}, vol.~5, no.~2, pp. 1672--1679, 2020.

\bibitem{saputra2021graph}
M.~R.~U. Saputra, C.~X. Lu, P.~P.~B. de~Gusmao, B.~Wang, A.~Markham, and
  N.~Trigoni, ``Graph-based thermal--inertial slam with probabilistic neural
  networks,'' \emph{IEEE Transactions on Robotics}, 2021.

\bibitem{jiang2022thermal}
J.~Jiang, X.~Chen, W.~Dai, Z.~Gao, and Y.~Zhang, ``Thermal-inertial slam for
  the environments with challenging illumination,'' \emph{IEEE Robotics and
  Automation Letters}, vol.~7, no.~4, pp. 8767--8774, 2022.

\bibitem{detone2018superpoint}
D.~DeTone, T.~Malisiewicz, and A.~Rabinovich, ``Superpoint: Self-supervised
  interest point detection and description,'' in \emph{Proceedings of the IEEE
  conference on computer vision and pattern recognition workshops}, 2018, pp.
  224--236.

\bibitem{jeong2006visual}
W.~Y. Jeong and K.~M. Lee, ``Visual slam with line and corner features,'' in
  \emph{2006 IEEE/RSJ international conference on intelligent robots and
  systems}.\hskip 1em plus 0.5em minus 0.4em\relax IEEE, 2006, pp. 2570--2575.

\bibitem{fu2020pl}
Q.~Fu, J.~Wang, H.~Yu, I.~Ali, F.~Guo, Y.~He, and H.~Zhang, ``Pl-vins:
  Real-time monocular visual-inertial slam with point and line features,''
  \emph{arXiv preprint arXiv:2009.07462}, 2020.

\bibitem{zhou2019ground}
D.~Zhou, Y.~Dai, and H.~Li, ``Ground-plane-based absolute scale estimation for
  monocular visual odometry,'' \emph{IEEE Transactions on Intelligent
  Transportation Systems}, vol.~21, no.~2, pp. 791--802, 2019.

\bibitem{chen2022eil}
W.~Chen, Y.~Wang, H.~Chen, and Y.~Liu, ``Eil-slam: Depth-enhanced edge-based
  infrared-lidar slam,'' \emph{Journal of Field Robotics}, vol.~39, no.~2, pp.
  117--130, 2022.

\bibitem{tarrio2015realtime}
J.~J. Tarrio and S.~Pedre, ``Realtime edge-based visual odometry for a
  monocular camera,'' in \emph{Proceedings of the IEEE International Conference
  on Computer Vision}, 2015, pp. 702--710.

\bibitem{schenk2017robust}
F.~Schenk and F.~Fraundorfer, ``Robust edge-based visual odometry using
  machine-learned edges,'' in \emph{2017 IEEE/RSJ International Conference on
  Intelligent Robots and Systems (IROS)}.\hskip 1em plus 0.5em minus
  0.4em\relax IEEE, 2017, pp. 1297--1304.

\bibitem{schenk2019reslam}
------, ``Reslam: A real-time robust edge-based slam system,'' in \emph{2019
  International Conference on Robotics and Automation (ICRA)}.\hskip 1em plus
  0.5em minus 0.4em\relax IEEE, 2019, pp. 154--160.

\bibitem{zhou2018canny}
Y.~Zhou, H.~Li, and L.~Kneip, ``Canny-vo: Visual odometry with rgb-d cameras
  based on geometric 3-d--2-d edge alignment,'' \emph{IEEE Transactions on
  Robotics}, vol.~35, no.~1, pp. 184--199, 2018.

\bibitem{forster2015manifold}
C.~Forster, L.~Carlone, F.~Dellaert, and D.~Scaramuzza, ``On-manifold
  preintegration theory for fast and accurate visual-inertial navigation,''
  \emph{IEEE Transactions on Robotics}, pp. 1--18, 2015.

\bibitem{theoryOfEdgeDetection}
C.~Bhabatosh \emph{et~al.}, \emph{Digital image processing and analysis}.\hskip
  1em plus 0.5em minus 0.4em\relax PHI Learning Pvt. Ltd., 2011.

\bibitem{campos2021orb}
C.~Campos, R.~Elvira, J.~J.~G. Rodr{\'\i}guez, J.~M. Montiel, and J.~D.
  Tard{\'o}s, ``Orb-slam3: An accurate open-source library for visual,
  visual--inertial, and multimap slam,'' \emph{IEEE Transactions on Robotics},
  vol.~37, no.~6, pp. 1874--1890, 2021.

\bibitem{Agarwal_Ceres_Solver_2022}
\BIBentryALTinterwordspacing
S.~Agarwal, K.~Mierle, and T.~C.~S. Team, ``{Ceres Solver},'' 3 2022. [Online].
  Available: \url{https://github.com/ceres-solver/ceres-solver}
\BIBentrySTDinterwordspacing

\bibitem{bloesch2017iterated}
M.~Bloesch, M.~Burri, S.~Omari, M.~Hutter, and R.~Siegwart, ``Iterated extended
  kalman filter based visual-inertial odometry using direct photometric
  feedback,'' \emph{The International Journal of Robotics Research}, vol.~36,
  no.~10, pp. 1053--1072, 2017.

\bibitem{xu2022fast}
W.~Xu, Y.~Cai, D.~He, J.~Lin, and F.~Zhang, ``Fast-lio2: Fast direct
  lidar-inertial odometry,'' \emph{IEEE Transactions on Robotics}, 2022.

\end{thebibliography}




\end{document}